\documentclass{article}

\PassOptionsToPackage{numbers, compress}{natbib}

\usepackage[final]{nips_2018}

\usepackage[utf8]{inputenc} 
\usepackage[T1]{fontenc}    
\usepackage{hyperref}       
\usepackage{url}            
\usepackage{booktabs}       
\usepackage{amsfonts}       
\usepackage{nicefrac}       
\usepackage{microtype}      

\usepackage{verbatim}

\usepackage{threeparttable}

\usepackage{graphicx}

\usepackage{enumitem}

\title{Transformer to CNN: Label-scarce distillation for efficient text classification}

\author{
  Yew Ken Chia\\
  Red Dragon AI\\
  Singapore\\
  \texttt{ken@reddragon.ai} \\
  \And
  Sam Witteveen\\
  Red Dragon AI\\
  Singapore\\
  \texttt{sam@reddragon.ai} \\
  \And
  Martin Andrews\\
  Red Dragon AI\\
  Singapore\\
  \texttt{martin@reddragon.ai} \\
}

\begin{document}

\maketitle

\begin{abstract}


Significant advances have been made in Natural Language Processing (NLP) modelling since the beginning of 2018.
The new approaches allow for accurate results, even when there is little labelled data, 
because these NLP models can benefit from training on both task-agnostic and task-specific unlabelled data.
However, these advantages come with significant size and computational costs.



This workshop paper outlines how our proposed convolutional student architecture, 
having been trained by a distillation process from a 
large-scale model,
can achieve $300\times$ inference speedup and $39\times$ reduction in parameter count.
In some cases, the student model performance surpasses its teacher on the studied tasks.



\end{abstract}

\section{Introduction}






The last year has seen several major advances in NLP modelling, stemming from previous innovations in 
embeddings  \cite{mikolov2013distributed} \cite{pennington2014glove} \cite{qi2018and} and 
attention models \cite{vaswani2017attention} \cite{elbayad2018pervasive} \cite{yang2016hierarchical}
that allow Language Models (LMs) to be trained on very large corpuses : 
For instance ELMo \cite{peters2018deep}, OpenAI Transformer \cite{radford2018language} and recently BERT \cite{devlin2018bert}.


In addition, the power of building on LM-enhanced contextualised embeddings,
using a fine-tuning approach on task-specific unlabelled data \cite{howard2018universal},
has shown huge benefits for downstream tasks (such as text classification) - 
especially in a typical industrial setting where labelled data is scarce.




In order to make use of these advances, this work shows how a model distillation process
\cite{hinton2015distilling}
can be used to train a novel `student' CNN structure from a much larger `teacher' Language Model.
The teacher model can be fine-tuned on the specific task at hand, 
using both unlabelled data, and the (small number of) labelled training examples available.
%
%
%
The student network can then be trained using both labelled and unlabelled data, 
in a process akin to pseudo-labelling \cite{lee2013pseudo} \cite{clark2018semi}.


Our results show it is possible to achieve similar performance to (and surpass in some cases) 
large attention-based models 
with a novel, highly efficient student model with only convolutional layers.

\begin{figure}[ht]
  \centering
  \includegraphics[width=1.0\textwidth]{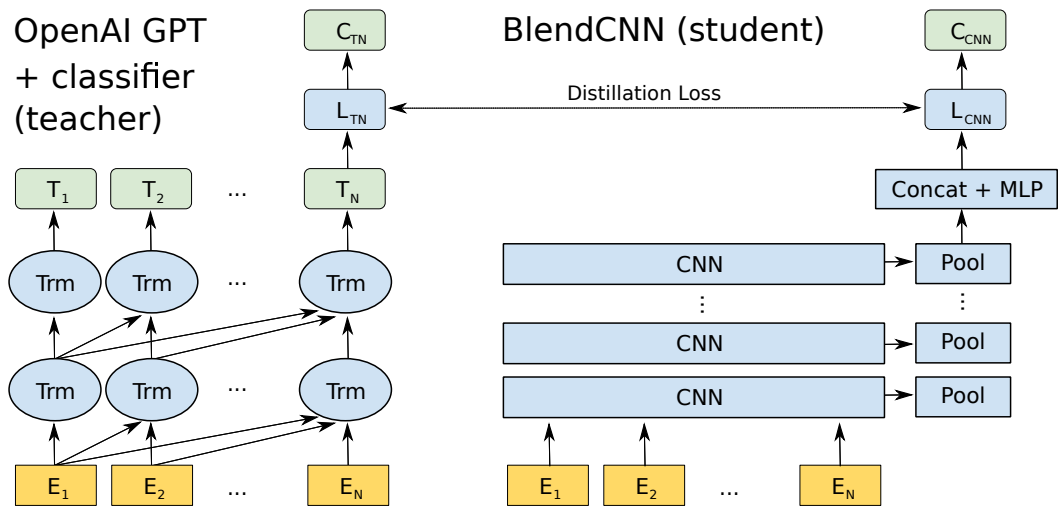}
    \caption{
      Model architecture with distillation across logits
    }
    \label{fig:architecture}
\end{figure}

\section{Model distillation}


In this work, we used the OpenAI Transformer \cite{radford2018language} model as the `teacher' 
in a model-distillation setting, with a variety of different `student' networks (see Figure \ref{fig:architecture}).

The OpenAI Transformer model consists of a Byte-Pair Encoded subword \cite{sennrich2015neural} embedding layer 
followed by 12-layers of ``decoder-only transformer with masked self-attention heads'' \cite{vaswani2017attention}, 
pretrained on the standard language modelling objective on a corpus of 7000 books.
This LM's final layer outputs were then coupled with classification modules 
and the entire model was discriminatively fine-tuned with an auxiliary language modelling objective, 
achieving excellent performance on various NLP tasks.



To optimize for speed and memory constraints of industrial deployment, a variety of
different models were trained (a)~on the classification task directly; 
and (b)~via distillation \cite{hinton2015distilling} of the logit layer output by the pretrained OpenAI classification model.


To combat label-scarcity and improve distillation quality, 
we inferred distillation logits for unlabelled samples in a pseudo-labelling
manner \cite{lee2013pseudo} \cite{clark2018semi}, 
while using transfer learning through pretrained GloVe embeddings \cite{pennington2014glove}.


\subsection*{Student models}

A number of common network structures were tested in the student role, specifically:
\vspace{-\topsep}
\begin{itemize}

\item a two-layer BiLSTM network \cite{liu2016recurrent}


\item a wide-but shallow CNN network \cite{kim2014convolutional}



\item a novel CNN structure, dubbed here `BlendCNN'
      


\end{itemize}

The BlendCNN architecture was inspired by the ELMo `something from every layer' paradigm, 
and aims to be capable of leveraging hierarchical representations for 
text classification \cite{yang2016hierarchical} \cite{yosinski2014transferable}.

The BlendCNN model is illustrated in Figure \ref{fig:architecture}, 
and comprises a number of CNN layers 
(with \verb+n_channels+=100, \verb+kernel_width+=5, \verb+activation+=\verb+relu+), 
each of which exposes a global pooling output as a `branch'.  
These branches are then concatenated together and ``blended'' through a dense network (\verb+width+=100), 
followed by the usual classification logits layer.  

\begin{table}[hb]
  \vspace{\topsep}
  \centering
  \caption{Parameter counts and inference timing}
  \label{efficiency-table}
  \begin{minipage}{.5\textwidth}
    \begin{tabular}{lrr}
      \toprule
   & Total & Sentences \\ 
   & parameters\textsuperscript{2} & per second\textsuperscript{3}  \\ 
      \midrule

2-Layer BiLSTM\textsuperscript{1} &    2,406,114  &  173.01   \\ 
KimCNN\tnote{1}         &    \textbf{2,124,824}  & 3154.57   \\
      \midrule
OpenAI Transformer      &  116,534,790  &  11.76    \\
8-layer BlendCNN        &    3,617,426  & 2392.34   \\
3-layer BlendCNN        &    2,975,236  & \textbf{3676.47}   \\ 

      \bottomrule
    \end{tabular}
  \end{minipage}%
  \begin{minipage}{.5\textwidth}%
    \begin{itemize}[noitemsep,topsep=0pt,parsep=0pt,partopsep=0pt]
      \footnotesize
      \item[1] BiLSTM and KimCNN model scores were lower
      \item[2] Parameter count estimate using a \verb+vocab_size+ of 20000 words, 
           except for OpenAI Transformer (which uses a byte-pair encoding embedding)
      \item[3] Timing measurement used \verb+n_samples+=1000, \verb+batch_size+=32, 
           based on actual time taken for K-80 GPU implementations
    \end{itemize}
  \end{minipage}%
\end{table}



\section{Experiments}





Each of the models was trained and tested on the 3 standard datasets described in \cite{P18-1216}
: AG News, DBpedia and Yahoo Answers.
The experiment proceeded in two phases, 
the first being to evaluate the performance of two baseline methods 
(TFIDF+SVM \cite{joachims1998text} and fastText \cite{grave2017bag})
along with that of the student networks (without the benefit of a LM teacher),
and the large LM, with a classification `head' trained on the task.
The second phase used the large LM in a `teacher' role, to train the other networks
as students via distillation of the LM classifier logits layer (with a Mean Absolute Error loss function).  

\begin{table}[t]
 \centering
 \begin{threeparttable}
   \caption{Scores for standard datasets}
   \label{scores-table}
   \begin{tabular}{llrlrlrl}
     \toprule
&   & \multicolumn{2}{c}{AG News}  & \multicolumn{2}{c}{DBpedia} & \multicolumn{2}{c}{Yahoo Answers} \\
    \midrule
\multicolumn{8}{l}{\small \textsc{Trained on 100 labelled examples per class}} \\ 
&    TFIDF + SVM                                 &  81.9 && 94.1 && 54.5  &\\
&    fastText                                    &  75.2 && 91.0 && 44.9  &\\
&    8-Layer BlendCNN                            &  87.6 && 94.6 && 58.3  &\\
&    OpenAI Transformer                          &  88.7 && 97.5 && 70.4  &\\
    \midrule
\multicolumn{8}{l}{\small \textsc{Trained by distillation\tnote{1} ~of OpenAI Transformer}} \\ 
&    2-Layer BiLSTM                              &  \textbf{91.2} && 97.0 && 70.5 &\\
&    KimCNN                                      &  90.9 && 97.6 && 70.4 &\\
&    3-Layer BlendCNN                           &  \textbf{91.2} &/ {\small 88.4\tnote{2}} & 98.2 &/ {\small 95.5} & \textbf{71.0} &/ {\small 63.4} \\
&    8-Layer BlendCNN                           &  \textbf{91.2} &/ {\small 89.9} & \textbf{98.5} &/ {\small 96.0} & 70.8 &/ {\small 63.4} \\
    \bottomrule
  \end{tabular}%
\begin{tablenotes}  
      \small
      \item[1] Distillation training used 100 labelled examples per class,  
               plus 10 times as many unlabelled examples as pseudo-labelled by the OpenAI LM
      \item[2] Small figures are results where distillation was conducted without unlabelled data
      \item[*] All CNNs use 100-dimensional trainable GloVe embeddings as input
      \item[*] Adam optimisation was used, with a constant learning rate of $10^{-3}$
\end{tablenotes}

 \end{threeparttable}
\end{table}

\section{Results}

Referring to Table \ref{scores-table}, 
the 3-Layer and 8-Layer variants of the proposed BlendCNN architecture 
achieve the top scores across all studied datasets.
However, the performance of the proposed architecture is lower
without the `guidance' of the teacher teacher logits during training, 
implying the marked improvement is due to distillation.
The additional results given for BlendCNN 
quantifies the advantage of adding unlabelled data into the distillation phase of the student model training.

Notably from Table \ref{efficiency-table}, the 3-Layer BlendCNN student has $39\times$ fewer parameters 
and performs inference $300\times$ faster than the OpenAI Transformer 
which it empirically out-scores.

\section{Discussion}

For text classifications, 
mastery may require both high-level concepts gleaned from language under
standing and fine-grained textual features such as key phrases. 
Similar to the larval-adult form analogy made in \cite{hinton2015distilling}, 
high-capacity models with task-agnostic pre-training may be well-suited 
for task mastery on small datasets (which are common in industry). 
On the other hand, convolutional student architectures may be more ideal 
for practical applications by taking advantage of massively parallel computation 
and a significantly reduced memory footprint.


Our results suggest that the proposed BlendCNN architecture can efficiently achieve higher scores 
on text classification tasks due to the direct leveraging of hierarchical representations,
which are learnable (even in a label-sparse setting) from a strong teaching model.




Further development of specialized student architectures 
could similarly surpass teacher performance 
if appropriately designed to leverage 
the knowledge gained from a pretrained, task-agnostic teacher model 
whilst optimizing for task-specific constraints.

\newpage

\bibliographystyle{unsrtnat}
\bibliography{../nips2018_all}

\end{document}